\documentclass[conference]{IEEEtran}
\IEEEoverridecommandlockouts
\usepackage{cite}
\usepackage{amsmath,amssymb,amsfonts}
\usepackage{algorithmic}
\usepackage{graphicx}
\usepackage{textcomp}
\usepackage{xcolor}
\usepackage{stfloats}

\usepackage{amsmath}
\usepackage{mathtools}
\usepackage{amssymb}
\usepackage{float}
\usepackage{bm}
\usepackage{mathdots}
\usepackage{subfigure}
\usepackage{hyperref}
\usepackage{enumitem}   
\usepackage{balance}

\def\BibTeX{{\rm B\kern-.05em{\sc i\kern-.025em b}\kern-.08em
    T\kern-.1667em\lower.7ex\hbox{E}\kern-.125emX}}

\begin{document}
\title{Example When Local Optimal Policies Contain Unstable Control}

\author{\IEEEauthorblockN{ Bing Song}
\IEEEauthorblockA{\textit{HP-NTU Digital}\\ \textit{Manufacturing Corporate Lab} \\
Singapore,  Singapore 639798 \\
bing.song@ntu.edu.sg}
\and
\IEEEauthorblockN{Jean-Jacques Slotine}
\IEEEauthorblockA{\textit{Nonlinear Systems Laboratory} \\
\textit{Massachusetts Institute of Technology}\\
Cambridge MA 02139-4307   \\
jjs@mit.edu}
\and
\IEEEauthorblockN{Quang-Cuong Pham}
\IEEEauthorblockA{\textit{ HP-NTU Digital}\\ \textit{Manufacturing Corporate Lab} \\
Singapore,  Singapore 639798 \\
cuong@ntu.edu.sg}
}

\maketitle

\newcommand{\TODO}[1]{{\color{red} {\bf #1}}}

\begin{abstract}
We provide a new perspective to understand why reinforcement learning (RL) struggles with robustness and generalization. We show, by examples, that local optimal policies may contain unstable control for some dynamic parameters and overfitting to such instabilities can deteriorate robustness and generalization. Contraction analysis of neural control reveals that there exists boundaries between stable and unstable control with respect to the input gradients of control networks. Ignoring those stability boundaries, learning agents may label the actions that cause instabilities for some dynamic parameters as high value actions if those actions can improve the expected return. The small fraction of such instabilities may not cause attention in the empirical studies, a hidden risk for real-world applications. Those instabilities can manifest themselves via overfitting, leading to failures in robustness and generalization. We propose stability constraints and terminal constraints to solve this issue, demonstrated with a proximal policy optimization example. 
\end{abstract}


\section{Introduction}
Reinforcement learning (RL) struggles with robustness and generalization when applying to continuous control problems. It is believed that overfitting to specific training environments causes those failures\cite{moos2022robust, xu2022trustworthy}. Here we provide another possibility. We show, by examples, that local optimal policies can contain unstable control for some dynamics. Overfitting to such instabilities instead of specific training environments can also deteriorate generalization.

\begin{figure}[t!]
    \centering 
    \subfigure[Intuition (tasks $f_j\in \mathbb F =\{f_1, f_2, f_3\}$).]{\includegraphics[width=3.3in]{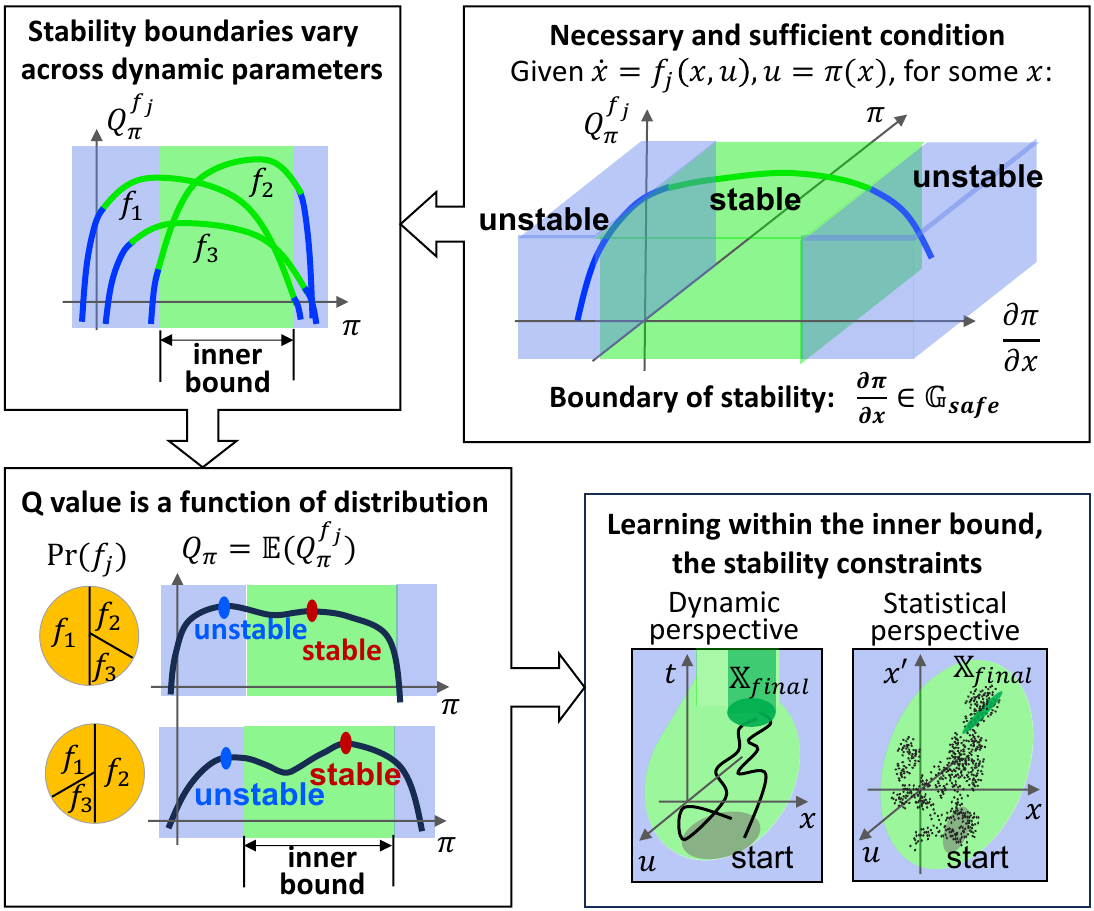}\label{fig:intuition}} \\
     \subfigure[PPO vs C(onstrained)-PPO.]{\includegraphics[width=3.3in]{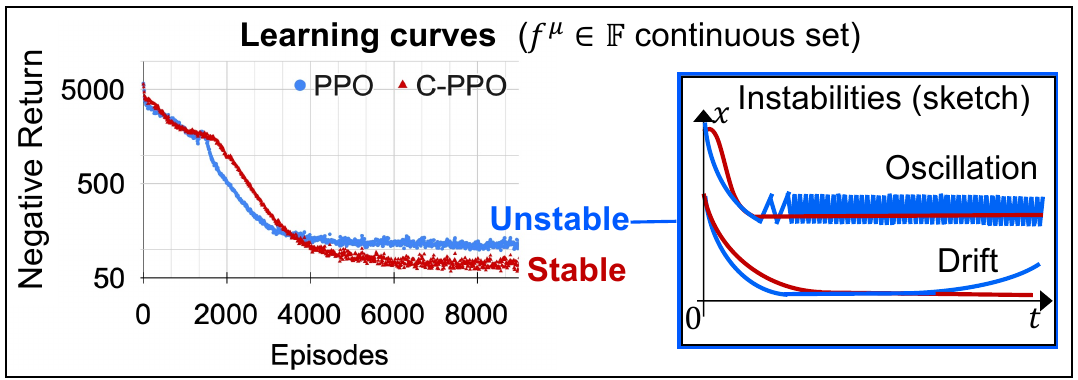}\label{fig:ppo}} 
    \caption{Optimality does not imply stability. (a) Local optimal policies can contain unstable control because the stability boundaries may not align with the landscape of expected returns. (b) A PPO example shows that local optimal policies introduce oscillation and drift if without stability constraints.}\label{fig:1}
\end{figure}

Local optimal policies may perform unstable control for some dynamic parameters as long as such unstable control can achieve higher expected returns for the task distribution, an example that shows optimality does not imply stability. In particular, for each set of dynamic parameters, there exists boundaries between stable and unstable control, which do not necessarily align with the reward (expected returns) landscape, because (i) there is no mathematical guarantee that optimal actions are stable unless value functions employed as Lyapunov functions and the like, and (ii) the reward landscape is a function of the parameter distribution while the boundaries are functions of the parameters regardless of their distribution. As the stability boundaries and the reward landscapes are not aligned, it allows (local) optimal policies to introduce instabilities into some dynamics. 

Such instabilities may be a small fraction of all possible dynamics, likely causing little attention in empirical studies, a hidden risk for real-world applications. Overfitting to such unstable control can corrupt learning, one possible culprit behind RL's failures in generalization.

It is difficult to leverage reward-based methods to prevent such instabilities. Sometimes it is even infeasible to rely on the reward landscape to capture all possible instabilities via reward engineering, because reward engineering requires predicting the form and magnitude of such instabilities from human experiences for the RL settings, which may involve large variations, uncertainties, and unknowns. The neural networks used in control policies further complicate the dynamics, challenging human predictions.

In this paper, we present the necessary and sufficient condition for stability via contraction theory, which reveals that there exist stability boundaries for each set of dynamic parameters, an inequality condition in terms of the input gradients of control networks (the network Jacobians). We show the differences between stability boundaries and reward landscapes with simple examples. 

To prevent unstable control from local optimal policies, we propose the stability constraints in \cite{song2023modularity} and the terminal conditions on system equilibria. We derive the method to force the system equilibrium at the goal, that is, feeding accumulative errors as inputs to control networks.  

We illustrate the issue and the solution with a proximal policy optimization (PPO) example. In the example, PPO converges to local optimal policies that create oscillation and drift, the amount of which is less than half a percent in testing, barely causing attentions without close examination. Those instabilities manifest themselves via overfitting in generalization and learning with large networks. Adding stability constraints and terminal conditions has prevented such instabilities and improved generalization dramatically. We also observe that feeding the accumulative errors to control networks improves the returns by an order of magnitude.

\subsection{Contributions}

The key contribution of this paper is to issue the warning of learning tasks via trial and error regardless of stability analysis. We point out that (i) local optimal policies may contain unstable control, as learning agents optimize neural control over task distributions while ignoring the stability boundaries, and (ii) such instabilities may be a small fraction but can manifest themselves through overfitting in generalization. To our best knowledge, this is the first time this phenomenon being illustrated in RL.

Other technical contributions include (i) the necessary and sufficient condition with respect to the input gradients of control networks for dynamic convergence via contraction theory, and (ii) feeding accumulative errors as inputs to control networks to force system equilibriums at goals.

We acknowledge that our analysis and examples are from the perspective of control theory. How often this phenomenon may happen in RL applications remains an open question.

\subsection{Related work}
Overfitting to some specific training environments is known to be the culprit in damaging the robustness and generalization of RL\cite{moos2022robust}. Here we identify another culprit, overfitting to the unstable control from local optimal policies, one distinct concern when learning to control via trial and error. 

Given a physical dynamic system represented as $\mathbf{\dot x} = \mathbf f^{\bm \mu}(\mathbf x, \mathbf u)$ where $\bm\mu$ denotes the parameters, RL optimizes the rewards over $\bm\mu\in \mathbb{M}$, the set of possible dynamic parameters.

The variations in $\bm\mu\in\mathbb{M}$ in RL settings are likely much larger than $\bm\mu  +\bm\delta_\mu$ in robust control or $\bm\mu(t)$ in adaptive control, which raise the concerns we demonstrated in this paper.

\subsubsection{Robustness of RL} 

One major direction is to formulate the robustness problem into robust Markov decision process (MDP) and seek the minimax policy\cite{lim2013reinforcement}, for example, adversarial RL\cite{lutjens2020certified, pinto2017robust, mandlekar2017adversarially}. Other directions include maximum entropy\cite{eysenbach2021maximum}, domain randomization\cite{tobin2017domain}, and risk-based RL\cite{mihatsch2002risk, pan2019risk}. Whether those methods can handle the issue depends on whether the unstable control can be exposed as unwanted behaviors by introducing adversaries and variations into training or whether the worst cases and risks can capture such instabilities. 

Another large group is to integrate control techniques with RL algorithms, leveraging control theory for control priors and regularization\cite{johannink2019residual, cheng2019control}. For example, Reference \cite{cheng2019control} improves robustness by providing control priors and applying regularization to reduce variances between updates, which encourages the learned policies near the control priors. Whether this kind of soft constraints can alleviate this issue is on a case-by-case basis. In our PPO experiment, the policies at early updates are stable and the learning curves imply small variances in returns, while instabilities still happen as learning progresses, of a very small amount. 

Along the line of integrating control with RL, stability certificates and constraints directly deal with control stability, mainly developed for safety-critical applications rather than for robustness and generalization. This paper illustrates how the three properties, control stability, robustness, and generalization, are interwoven. 

The following discusses the related work in the broader topic: the safety of RL. 

\subsubsection{Safety of RL}

Safe RL includes many research lines to mitigate different safety concerns and one may find the details in the review paper \cite{brunke2022safe}. 

Instead of collision avoidance in navigation, the safety concern here is the stability of control. In particular, we observe small amounts of oscillation and drift in the example that may cause extra tool wear, possibly leading to accidents in practice. 

Besides the soft constraints via control priors and regularization, many studies try to introduce the results of neural certificates from nonlinear control into RL\cite{chang2021stabilizing, ma2022joint}. Neural certificates based on Lyapunov theory, control barrier functions, contraction theory have achieved wide success in the field of nonlinear control. Reference \cite{dawson2022safe} provide a survey of the Lyapunov-based certificates and Reference \cite{manchester2021contraction} presents a tutorial overview of contraction-based methods in learning. To introduce those results into RL is intrinsically difficult. The success of neural certificates depends on the shrinking bounds of the generalization error of those learned certificates. Provided the control policy is simultaneously updated by trial and error for rewards, the possible conflicts between the generalization of stability certificates and the exploration for maximizing rewards. It is still lacking the theoretical analysis of the interplay between generalization of certification and exploration for rewards. One resulting issue is being unnecessarily conservative. 

Leveraging dynamics models, in particular, employing confidence intervals that evaluates how good the measures represent the true function, Reference \cite{berkenkamp2017safe} can realize both stability and full exploration, under the regularity assumption of a Gaussian process prior. This method relies on that Gaussian process can be use as a model of the dynamics. 

Another line is to limit the input gradients of control networks. Both in this work and in \cite{jin2020stability}, it is proved that the necessary condition for stability yields constraints on the input gradients of control networks. In implementation, it is difficult to derive the necessary and sufficient bounds and sufficient bounds can be applied as substitutes, implying conservatism. To alleviate such conservatism,\cite{jin2020stability} adjusts the boundary based on the uncertainties; and our work \cite{song2023modularity} separates control and planning via hierarchical learning to allow free exploration in planning while adding constraints on control for stable and compliant response to contacts. 

\subsubsection{Generalization of RL}

Characterizing generalization is an on-going topic in RL\cite{packer2018assessing} and we noticed that different assumptions about transition functions may be used in measuring generalization\cite{witty2021measuring}. We remark that the phenomenon we show here, the highly rewarded unstable control, results from the varying transition function that depends on the dynamic parameter distribution, a valid concern for real world applications to learn similar tasks. Our task settings belong to zero-shot generalization\cite{2021arXiv211109794K}. 

Efforts from different research lines try to analyze why RL struggles with generalization. For example, Reference \cite{song2019observational} analyzes the overfitting from the perspective of supervised learning, overfitting to the observations that are irrelevant to the latent dynamics of the task family that shares the similar MDPs; Reference \cite{malik2021generalizable} analyzes the metrics to define the similarity of tasks, in particular, the structural conditions that allow efficient generalization; Reference \cite{igl2020transient} revisits the non-stationarity in RL and shows that changes in data distribution during training may damage generalization. Here we contribute from the perspective of control theory. We illustrate that optimality does not imply stability and overfitting to the small instabilities from local optimal policies can also deteriorate generalization.

\subsubsection{On the basis of control theory}

Our theoretical analysis is based on contraction theory\cite{lohmiller1998contraction} and optimal control\cite{kirk2004optimal}; adding integral to realize the terminal constraint is inspired by integral control that cancels the steady state error. The stability constraint used in the experiments is from \cite{song2023modularity}, a minimally invasive method to guarantee control stability for RL by leveraging modularity.

Oscillation or chattering is a phenomenon relative to the loss of Lipschitz continuity in control. The underlying reason may vary, for example, from unmodeled high frequency dynamics and digital controllers with finite sampling rate in sliding mode control\cite{lee2007chattering}. In the PPO example here, the oscillation is related to the small stability margins\cite{keel1998stability} and the optimization-based nature similar to \cite{6760327}. 

Drift is likely distinct for RL, enabled by the large capacities of control networks as discussed in Section.~\ref{goal_equilibrium}.

\subsection{Paper organization}
This paper is organized as follows: 
\begin{enumerate}
\item Problem formulation. 
\item Stability boundaries.
\item Terminal conditions
\item Optimality does not imply stability: PPO example.
\item Concluding remarks. 
\end{enumerate}

\section{Problem formulation}

Dynamics involved in the tasks can be represented as
\begin{equation}
\{\dot{\mathbf x} = \mathbf f(\mathbf x, \mathbf u), \mathbf f\in\mathbb{F}\}
\end{equation}
Denoting the coefficients of $\mathbf f$ by $\bm\mu$, 
\begin{equation}
\mathbb F =\{ \mathbf f^{\bm \mu}(\mathbf x, \mathbf u) | \bm \mu \in \mathbb M \}
\end{equation}
where $\mathbb M$ is the set of possible dynamic parameters, the distribution of which depends on the variety and uncertainties in the physical parameters of environments and robots. In this paper, we use $\mathbf f$ and $\mathbf f^\mu$ interchangeably. 

To handle similar tasks at the presence of uncertainties in the real world, the variations in $\bm\mu\in\mathbb M$ are much more complicated than the small model errors $\bm\delta_\mu$ in robust control or the gradual changes in $\bm\mu(t)$ for a plant in adaptive control. 

This model can be equivalently represented as state transition probability functions, given an initial state distribution and a sampling frequency. Note that the transition function varies along with the task parameter distribution.

Intuitively, learning to control the tasks can be viewed from the dynamic perspective by plotting $(x, u, t)$ and from the statistical perspective by plotting $(x, u, x')$, as illustrated in the bottom right sketch in Fig.~\ref{fig:intuition}.

%
%
%

\section{Stability boundaries}

\subsection{Necessary and sufficient condition}\label{condition}
Applying contraction theory to the neural control system, we can obtain the necessary and sufficient condition with respect to the input gradients of the networks (the Jacobian) that guarantees dynamic convergence at the presence of bounded disturbances and with bounded initial state distributions, regardless of the limit behavior of the nonlinear system, that is, without specifying equilibria, limit cycles, and the like.

{\underline{Necessary and sufficient condition for stability}:}\label{condition} 
Given the system $\dot{\mathbf x} =\mathbf f(\mathbf x, \mathbf u)$ controlled by $\mathbf u=\bm\pi(\mathbf x)$, where $\mathbf f$ and $\bm\pi$ are continuously differentiable, if there exists uniformly positive definite matrix $M(\mathbf x, t)$, $\forall \mathbf x, t$, s.t. $\exists \alpha \in \mathbb R^+$,$\forall \mathbf x, t$, 
\begin{equation}\label{constraint}
\dot M + M(\frac{\partial \mathbf f}{\partial \mathbf x}+\frac{\partial \mathbf f}{\partial \mathbf u}\frac{\partial \bm\pi}{\partial \mathbf x}) +(\frac{\partial \mathbf f}{\partial \mathbf x}+\frac{\partial \mathbf f}{\partial \mathbf u}\frac{\partial \bm\pi}{\partial \mathbf x})^TM\preceq -2\alpha M 
\end{equation}
the system is incrementally exponentially stable. The converse also holds.

The proof can be obtained directly by applying Theorem 2.1 in \cite{tsukamoto2021contraction} to the system $\dot{\mathbf x} =\mathbf f(\mathbf x, \bm\pi(\mathbf x))$. Theorem 2.1 is the fundamental theorem of contraction theory, the proof of which can be found in \cite{tsukamoto2021contraction, lohmiller1998contraction}. 

The necessary and sufficient condition shows that there exists a safe set of $\partial \bm\pi/\partial \mathbf x$ that satisfies Eq.~\eqref{constraint}, which defines the boundary between stable and unstable control. Similar results are obtained for $L_2$ stability in \cite{jin2020stability}.

\subsection{Stability boundaries vary across dynamic parameters.}

It is challenging and possibly infeasible to solve for the stability boundaries, while analytic solutions usually exist for linear systems. 

Stability boundaries change along with the task parameters. 
For example, given $\dot x=f^{\mu_1}(x, u)=-x+u$ and $\dot x=f^{\mu_2}(x, u)=-x+20u$, the boundaries are $\partial \pi/\partial x<1$ and $\partial \pi/\partial u <0.05$ respectively. 

Discretization or sampling also changes the boundary. Given a continuous system $\dot x = a_1 x + a_2 u$, where $a_1, a_2\in \mathbb R$, its stability boundary is 
\begin{equation}\label{linear}
(a_1+a_2 \frac{\partial \pi}{\partial x})<0
\end{equation}
Provided a sampling period $\Delta t$, the resulting discrete system can be represented as $x(k+1) = (1+\Delta t\cdot a_1) x(k) + \Delta t\cdot a_2 u(k)$ and its stability boundary becomes $
|1+\Delta t\cdot a_1 + \Delta t \cdot a_2\frac{\partial \pi}{\partial x}|<1
$,
yielding 
\begin{equation}\label{discrete_linear}
-\frac{2}{\Delta t}<(a_1+a_2 \frac{\partial \pi}{\partial x})<0
\end{equation}
Given the sampling frequency approaches infinity, the boundary for the sampled system Eq.~\eqref{discrete_linear} converges to the boundary of the continuous system Eq.~\eqref{linear}. 

In the above examples, the stability boundaries in terms of $\partial \pi/\partial x$ are varying along with $\bm\mu = \begin{bmatrix} a_1, a_2, \Delta t \end{bmatrix}$.

\subsection{Q value is a function of distribution}

We show an example for illustration. The set of two dynamics 
\begin{equation}
\Big\{x(k+1) = \bm\mu^T \begin{bmatrix}x(k) \\ u(k)\end{bmatrix}, \bm\mu\in \{ \begin{bmatrix}0.1 \\ 0.1\end{bmatrix}, \begin{bmatrix}0.1 \\ 0.11 \end{bmatrix}\}\Big \}
\end{equation}
yields $Pr(0.1| 0, 1)=Pr(\bm\mu_1)$ and 
$Pr(0.11| 0, 1)=Pr(\bm\mu_2)$. The Q-value at $x=0, u=1$ becomes 
\begin{equation}
Q(0,1)=Pr(\bm\mu_1)Q^{\bm\mu_1}(0,1) + Pr(\bm\mu_2)Q^{\bm\mu_2}(0,1)
\end{equation}

\section{terminal conditions}\label{goal_equilibrium}
Terminal conditions here refer to the ones in optimal control derived from the calculus of variations \cite{kirk2004optimal}, which formulating optimal control problems into nonlinear two-point (initial and terminal) boundary-value problems. 

Here we propose a method to satisfy the terminal conditions for the goal tasks without time constraints, that is, to force the system equilibrium at the goal. Otherwise, drift in Fig.~\ref{fig:ppo} is encouraged as learning agents can exploit the capacity of control networks to keep the dynamics temporarily near the goal for high rewards.

To enforce the desired goal to be the steady-state equilibrium, we include an integral in the RL state, $\mathbf s=[\mathbf e, \int \mathbf e dt]^T$, that is, feeding the accumulative error into the control networks. Intuitively, with the accumulative error as inputs to the control policies, the system won't reach its steady state unless the current error being zero. 

\underline{Theorem of system equilibrium}\label{equilibrium}: Given a nonlinear system $\dot {\mathbf e} =\mathbf f(\mathbf e, \mathbf u)$ with a neural control policy $\mathbf u=\bm\pi$, where $\mathbf e=\mathbf x_{goal}-\mathbf x$, the system allows one and only one equilibrium at the goal, provided that 
\begin{itemize}
 \item the neural control system is incrementally exponentially stable in a convex region,
\item the input to the control policy include the accumulative error and the current error, 
\begin{equation}\label{u}
\mathbf u =  \bm\pi( \begin{bmatrix} \mathbf s_1 \\ \mathbf s_2 \end{bmatrix})=\bm\pi( \begin{bmatrix} \mathbf e  \\ \int_0^t\mathbf e dt \end{bmatrix})
\end{equation}
\item $\partial \bm\pi/\partial \mathbf s_2$ is full rank, i.e., if and only if $\mathbf e=0$, $\frac{\partial \bm\pi}{\partial \mathbf s_2}\mathbf e =0$.
 \end{itemize}

Proof: At the equilibrium, $\dot{\mathbf e}=0$ yields $\mathbf f(\mathbf e, \bm\pi(\mathbf e)) =0$, which allows non-zero equilibrium, i.e., $\mathbf f(\mathbf e_{ss}, \bm\pi(\mathbf e_{ss})) =0$ where $\mathbf e_{ss}\neq 0$. 

This non-zero equilibrium encourages drift. If the dynamics for a given $\mathbf f$ converges in a convex region, there exists at most one equilibrium point, since any distance between two trajectories is shrinking exponentially in that region (Section 3.7 in \cite{lohmiller1998contraction}). When there exists one and only one equilibrium that is not at the goal, drifting is encouraged because (i) when $t\leq N$, the trajectory length used in training, deep RL tries to reach the goal for higher rewards and (ii) as $t\rightarrow \infty$, the dynamics will move to its equilibrium away from the goal.

To avoid such complications, we feed the integral of $\mathbf e$ and the current $\mathbf e$ as inputs to the network controller Eq.~\eqref{u}. This adds an extra condition $\dot{\mathbf u} =0$ for steady state, i.e., $
\dot{\mathbf u} = \frac{\partial \bm\pi}{\partial \mathbf s_1}\dot{\mathbf e}
 +\frac{\partial \bm\pi}{\partial \mathbf s_2}\mathbf e=0$. Then the system equilibrium should satisfy
\begin{equation}\label{zero_err}
\begin{cases}
\mathbf f(\mathbf e, \bm\pi(\mathbf e, \int \mathbf e dt))  =\mathbf {0}
\\[8pt]
\frac{\partial \bm\pi}{\partial \mathbf s_2}\mathbf e =0
\end{cases}
\end{equation}
Given a full rank $\partial \bm \pi/\partial \mathbf s_2$ and a control policy with enough capacity that $\mathbf f(\mathbf e, \bm\pi(\mathbf e, \int \mathbf e dt)) = \mathbf{0}$, there exists one and only one solution at $\mathbf e=0$. 

Proof is finished.

\section{Optimality does not imply stability:\\ PPO example}

Optimality does not imply stability, while it is not unusual in the history of optimal control that the theoretical studies in stability lag behind the empirical success of optimal control techniques. In Kalman's paper in 1960\cite{kalman1960contributions} that sets the theoretical foundation of optimal control via the calculus of variations, it states ``In the engineering literature it is often assumed (tacitly and incorrectly) that a system with optimal control low is necessarily stable." A recent example is model predictive control (MPC)\cite{mayne2014model}, whose successful adoption in process industries was years ahead of theoretical studies in its stability, as stability is always achieved via sufficiently long horizons in the process industries. The theoretical studies characterize the application scope and derive techniques to expand the scope. For example, \cite{grimm2004examples} shows that with discontinuities in the value function and control law in MPC, there exists zero robustness: an arbitrarily small disturbance can destabilize the control system.

A similar pattern is happening to RL. Here we contribute to the stability studies in RL from handling dynamic parameter distributions. 
This section shows the PPO example that local optimal policies introduce oscillation and drift into some trajectories. We apply stability constraints to mitigate such concerns and improve the performance.

This section is organized as follows: (i) peg-touching control task, (ii) remarks on the control problem, (iii) implementation of stability constraints, (iv) experiment 1: terminal constraints, and (v) experiment 2: stability constraints.

\subsection{Peg-touching control task}\label{cppo}

\subsubsection{Task settings}

The task is to move a 2D ``peg" to touch a surface at the desired position $x_d$ with the desired force $f_d$ along the z-axis as sketched in Fig.~\ref{fig:ppo_task}. This task is to approximate stiff robots with powerful motors that can quickly reach the goal, while contacts on different surfaces require complex controllers because of different stiffness, like in \cite{pham2020convex}, a 6-axis Denso VS-060 robot modeled as a first-order dynamics with time constant 0.0437 s. 

We assume the dynamics of the peg-robot as follows
\begin{equation}
\tau_x \dot x + x = u_x, \ \  \tau_z \dot z + z = u_z 
\end{equation}
where $\tau_x=0.0437$, $\tau_z=0.01$, $x\in[1, 3]$, and $z\in[-3, 1]$. Given the surface profile $g(x)=K_1\sin(x)+K_2\cos(x)$ with randomly sampled coefficient $K_1\in [-0.01, 0.01]$ and $K_2\in [-0.01, 0.01]$, the contact force can be approximated by 
\begin{equation}
f = K_{sur}\min{(z-g(x),\  0)} 
\end{equation}
where $K_{sur}$ represents the stiffness of the surface material, randomly sampled from $K_{sur}\in [1, 31]$. The desired $x_d$ and $f_d$ are also randomly sampled at the beginning of each trajectory.

For this peg-touching task, the task parameters are 
\begin{equation}\label{phi}
\bm\mu =\begin{bmatrix} \tau_x & \tau_z  & K_1 & K_2 & K_{sur} \end{bmatrix}
\end{equation}

We use PPO to learn the tasks by optimizing 
\begin{equation}\label{2d_dynamics}
\min  \mathbb{E} \Big (\sum_{k=0}^N \ \lVert \mathbf e(k) \rVert  + 0.5\lVert \mathbf{u}(k)-\mathbf u(k-1) \rVert\Big)
\end{equation}
where $N=8$ s in training (4000 steps per episode, 0.002 s per time step), given the peg is able to move to the close vicinity of the desired goal in less than 2 s. To close examine if drift happens, in testing $N=16$ s.

\subsubsection{Algorithm and configuration search} We apply PPO with fully connected layers (tanh as activation) for this task. We searched the best configuration for PPO without constraints, which is also used for PPO with stability constraints (C-PPO). The resulting best configuration is $\mathbf s=[\mathbf e,\Delta t \sum \mathbf e]^T$, network sizes $3\times 32$, and the hyperparameters are in Table.~\ref{tab:ppo} (otherwise the default parameters from the OpenAI baselines). Five seeds are tested and similar results are observed. 
    \begin{table}[h!]
    \caption{PPO Parameters}
    \centering
    \begin{tabular}{|c|c|}
    \hline
entropy coefficient  & 0. \\
\hline
learning rate & 0.0001 \\
\hline 
vf coefficient & 1.0 \\
\hline
max gradient norm & 0.5 \\
\hline 
discount factor & 0.99 \\
 \hline
 lambda & 0.95 \\
 \hline 
number of minibatches & 4. \\
\hline
clip range & 0.1 \\
\hline
number of epoches & 10  \\
 \hline 
    \end{tabular}
    \label{tab:ppo}
\end{table}

\subsection{Remarks on the control problem} 
This control task can be well performed with a simple feedback controller $\mathbf u=-K\mathbf e$, where $K$ is inside the stable range, $\mathbf u = [u_x, u_z]^T$, and the tracking error $\mathbf e=[x_d - x,  f_d -f]^T$. This simple feedback controller can move the peg to the desired goal with zero errors and without oscillation nor drift. 

However, to optimize the trajectories over the parameter distributions is complicated. Firstly, the fast response ($\tau_x=0.0437$ and $\tau_z=0.01$) implies a relatively small stability margin for $\partial \bm\pi /\partial \mathbf s$ given a sampling frequency. We use the sampling period 0.002 s (500 Hz) in the experiments. Secondly, contacts on the randomly created surface profile introduce unpredictable discontinuities into the first order derivatives, i.e., piecewise-smooth. Thirdly, the large range of parameters likely results in many local optima as well as further narrowing the inner bound of stability boundaries. 

Possible measures other than stability constraints to prevent oscillation and drift are as follows. 

Oscillation happens because learning agents optimize the expected rewards regardless of Lipschitz continuity in control, similar to the chatters in \cite{6760327}. To prevent oscillation, a penalty on the first derivative of actions is usually added in the reward, which also limits fast changes in motion. Balancing the trade-off from reward engineering is difficult for the tasks that prefer fast response at some states and accurate adjustments at some other states, while dynamic convergence guarantees can also discourage oscillation without limiting the fast response.

Drift happens because of finite horizons. When the system equilibrium is not at the goal, a learning agent could be capable of temporarily forcing the robot at the goal for rewards. But beyond the horizon, the robot will drift away. Theoretically, one can keep increasing the horizon used in training until drift disappears, which is not preferred in practice due to the extra time and financial costs. With constraints for dynamic convergence, drift can be prevented without the extra costs. 

Note that extra measures are already implemented in the task settings to avoid oscillation and drift: (i) To discourage oscillation, the penalty on $\lVert \mathbf{u}(k)-\mathbf u(k-1) \rVert^2$ is added in the step-wise reward. (ii) To discourage drift, the trajectory length $N$ in training is approximately 4 times of the length needed to reach any point in the task space. The experiment results are with the above measures.

\subsection{Implementation of stability constraints}\label{stability_constraint} In the necessary and sufficient condition for stability in \ref{condition}, the matrix $M(\mathbf x, t)$ represents a Riemannian metric, which defines how all possible trajectories converge. Provided a metric $M$, the stability constraints can be derived via Eq.~\eqref{constraint}. 

For any converging dynamics with bounded transients, there exists such a metric with Eq.~\eqref{constraint} being satisfied, that is, the necessary condition, while it is challenging to prove that given a metric, the actions beyond the stability constraints is unstable for all possible other metrics. To avoid such difficulties, stability constraints are usually sufficient conditions, which is the same case for Lyapunov based methods. The metric in contraction theory corresponds to a differential Lyapunov function (see the proof of Theorem 2.1 in \cite{tsukamoto2021contraction}).

It is also challenging to find a metric or derive the stability constraints from the nonlinear inequality equations. We apply the method in \cite{song2023modularity}, which leverages the modularity of contraction to simplify the problem, resulting in a minimally-invasive method for stable neural control. The details of the implementation for this experiments is in Appendix F in \cite{song2023modularity}

The stability constraints are derived with dynamic models. The robustness to model errors (the bounds on model errors within which the constraints are valid) can be derived via contraction theory, yielding implicit inequalities on the derivatives of the dynamics ($\partial \mathbf f/\partial \mathbf x$ and $\partial \mathbf f/\partial \mathbf u$), which is presented in Appendix E in \cite{song2023modularity}. 

In the generalization test, we used the nominal dynamic models from the parameter distribution in training that contain model errors to derive the stability constraints. By doing so, we can avoid extra information about the parameter distribution in testing provided to C-PPO and meanwhile test the robustness.

\subsection{Experiment: Terminal constraint}\label{soft_terminal}

The proposed method for terminal conditions, feeding the accumulative error into control networks, brings two changes: (i) new information to the learning agents, and (ii) constraints on system equilibrium. From the theorem in \ref{equilibrium}, to force the equilibrium at the goal includes two other conditions, $\partial \bm\pi/\partial \mathbf s_2\neq 0$ and incrementally exponentially stable in a convex region. 

Here we test the influence from using $\mathbf s=[\mathbf e,  \Delta t\sum\mathbf e]^T$ regardless the other two conditions, which can be seen as soft terminal constraint. In the next experiment for stability constraints, we close examine if drift can happen with the soft terminal constraint, comparing with the constrained PPO that the other two conditions are also satisfied. 

\begin{figure*}[ht]
\begin{minipage}[b]{0.26\linewidth}
\centering
\includegraphics[width=1.7in]{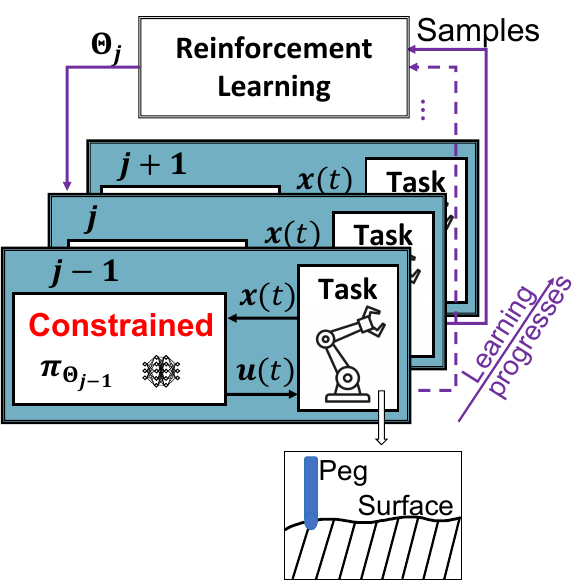}
\caption{RL with constraints for the peg-touching task, to reach a randomly sampled goal (force and position).}
\label{fig:ppo_task}
\end{minipage}
\ \
\begin{minipage}[b]{0.7\linewidth}
\centering
\includegraphics[width=5.in]{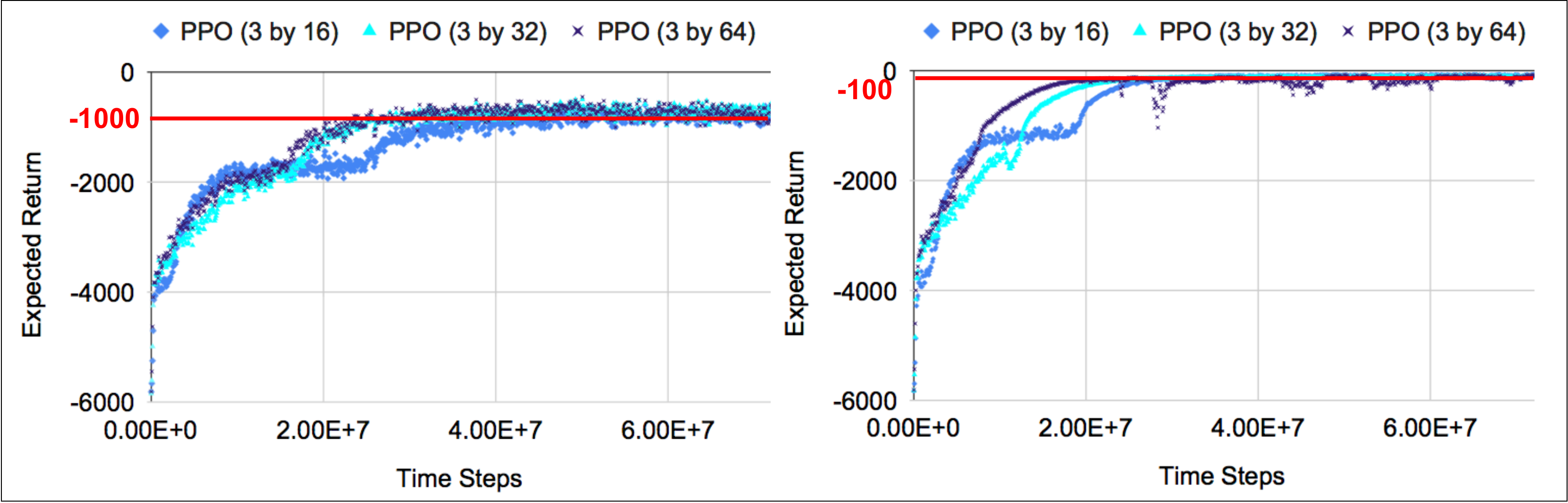}
\caption{Soft terminal constraint improves the rewards by an order of magnitude. Left: PPO. Right: PPO with $\mathbf s=[\mathbf e, \Delta t\sum \mathbf e]^T$, the soft terminal constraint that encourages the equilibrium of the neural control system at the goal. The expected return is improved from -1000 to -100.}
\label{fig:terminal_constraint}
\end{minipage}
\end{figure*}

Results are plotted in Fig.~\ref{fig:terminal_constraint}, the learning curves of network sizes $3\times 16$, $3\times 32$, and $3\times 64$. All learning curves with different network sizes are stable and converged to the similar level. Feeding the accumulative errors into the control networks, the expected return is improved by an order of magnitude, from -1000 (left) to -100 (right).

 \subsection{Experiment: Stability constraints}
 
 We test the influence from stability constraints via three aspects: (i) stability, (ii) generalization and robustness, and (iii) prevention of overfitting.
 PPO with the soft terminal constraint in Section~\ref{soft_terminal} is used as the baseline, denoted by PPO. 

\subsubsection{Stability test.} We closely examine if local optimal policies from PPO can introduce oscillation and drift into the dynamics. 

Given stable learning curves with similar levels of returns from PPO and C-PPO, $-100$ vs. $-70$ respectively in Fig.~\ref{fig:stability}, we tested 8000 trajectories every 1k episodes, starting from Episode 4000 until Episode 9000. 

Oscillation and drift are observed in PPO (Fig.~\ref{fig:stability}). We counted trajectories with error norms (averaged norm of the position and force errors from the final 5 steps) larger than 0.4, among which, most are oscillating, a fair amount drifting, and a few staying at the large steady state error with small drifting behaviors. 

The fraction of oscillation and drift is increasing as learning progresses (Fig.~\ref{fig:stability}), implying overfitting behind the scene. 

The fraction of such instabilities is small. The biggest fraction at Episode 9000 is less than half a percent, which can be easily ignored in empirical studies. Note that instabilities with small errors are not counted, since in practice such errors are likely covered by noises or sensor errors. If we count $\lVert \mathbf e\rVert>0.2$, in PPO, there exists $0.06\%$ at Episode 5000, $0.14\%$ at 6000, and $0.85\%$ at 9000. 

Adding stability constraints (C-PPO) can mitigate such concerns, the red line in Fig.~\ref{fig:stability} remaining at zero. 

Therefore, from learning curves and testing with the same task distributions, it seems that PPO and C-PPO achieve similar performance. However, close examine reveals that instabilities have been introduced into the local policies from PPO. Those instabilities manifest themselves in generalization and learning with large networks via overfitting in the following tests.

Note that although in this example that PPO has initially converged to stable suboptimal policies at Episode 4000, there is no guarantee that this always happens, just like that there is no guarantee that PPO always converges to the unstable local optima. The path of convergence depends on the task distribution, the initial distributions, the hyperparameters, etc.

\subsubsection{Generalization and robustness test.} 
We further test the policies at Episode 5000 for generalization and robustness, the policies that have $\sim 0.01\%$ instabilities in the stability test using the same task parameter distributions.  

We create the new task distribution by adding randomly sampled shift values $\Delta\bm \mu$ to the nominal task parameters $\bm\mu^{n}$ that are sample from the parameter distribution used in training. Recall that the parameters for this peg-touching task $\bm\mu$ in Eq.~\eqref{phi} includes $\tau_i$, $K_j$, and $K_{sur}$, where $i=x,z$ and $j=1,2$. The task parameter used in testing can be represented as $\bm\mu^{test}= \bm\mu^{n}+\Delta\bm \mu$
where $\Delta \bm\mu$ is sampled from the uniform distributions within $\pm 50\%$ $\tau_{i}^{n}$, $\pm 50\%$ $K_{j}^n$, and $\pm 100\%$ $K_{sur}^{n}$, with the constraint on the surface stiffness $K_{sur}^{test}>1$.

Note that the stability constraints are derived with the nominal task parameters $\bm\mu^{n}$. In other words, the model errors in the model-based constraints are equal to the $\Delta\bm \mu$.

The histograms of errors (average error from the final 5 steps) from 8000 randomly sampled trials are plotted in Fig.~\ref{fig:robustness}. Errors larger than 0.1 (force) and 0.01 (position) are counted. 

Most trajectories from PPO can not achieve the tracking accuracy for the new task distribution, with the largest tracking error around  2.5 (force) and 0.25 (position), much worse than in the stability test with the same task distribution (around $0.01\%$, that is, 8 trajectories with the norm of force and position larger than 0.4). 

C-PPO maintains the tracking accuracy on the similar level, with $0.01\%$ (8 out of 8000) with force errors around 0.14 and position error around 0.008.

\subsubsection{Prevention of overfitting}
\begin{figure*}[ht!]
 \centering
\subfigure[Stability: oscillation and drift from local optimal policies.]{{\includegraphics[width=3.83in]{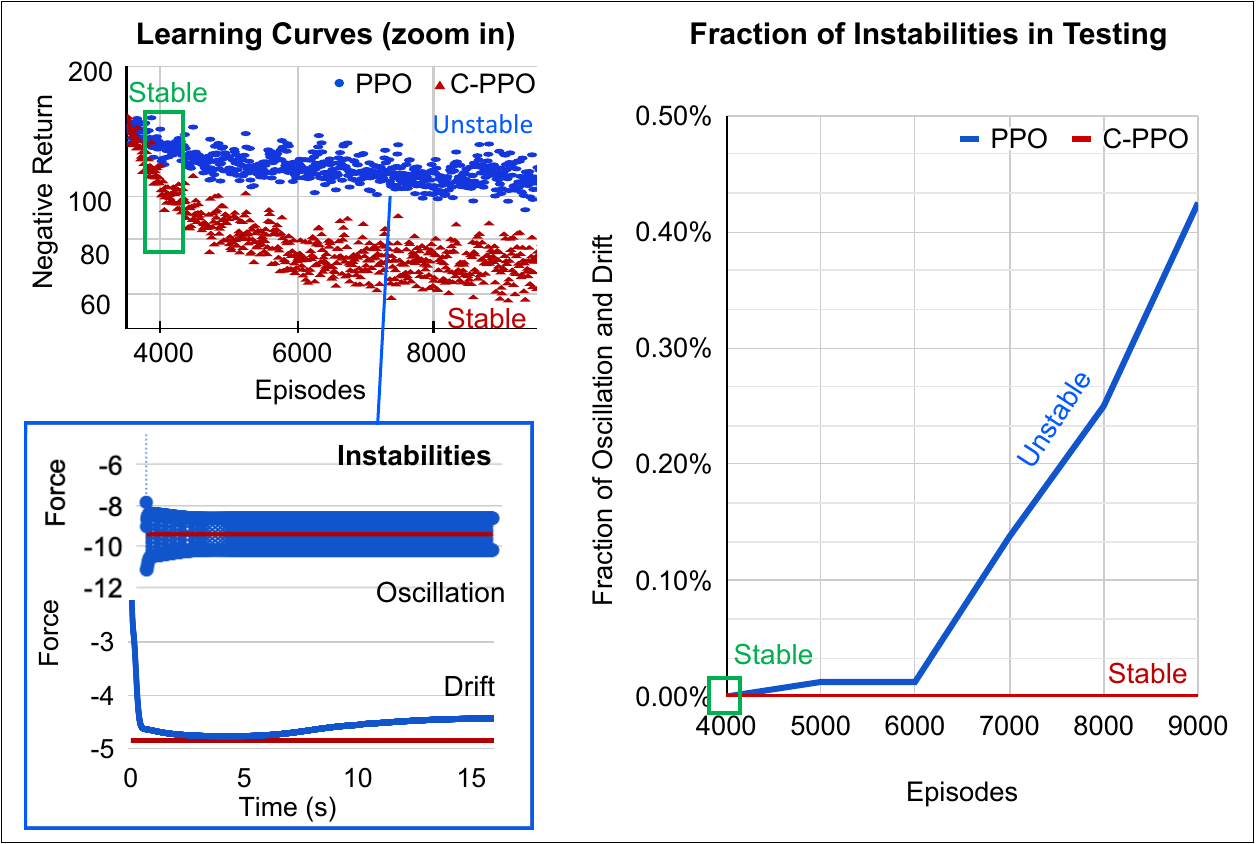}}\label{fig:stability}} 
    \subfigure[Generalization and robustness.]{{\includegraphics[width=1.59in]{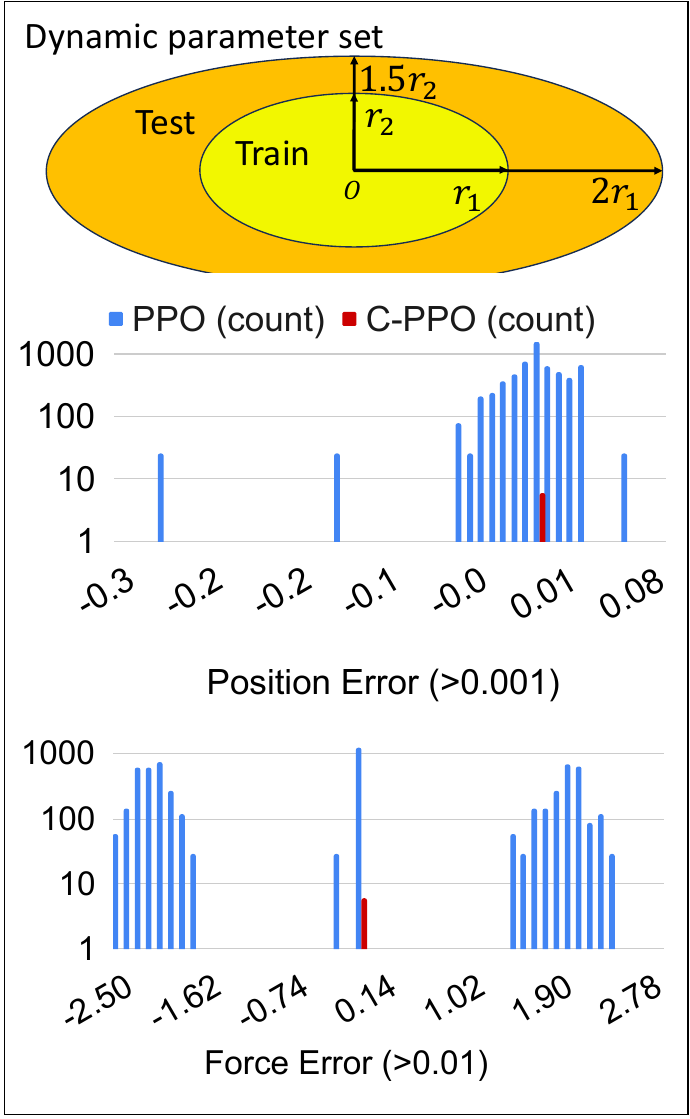}}\label{fig:robustness}}
       \subfigure[Prevention of overfitting.]{{\includegraphics[width=1.5in]{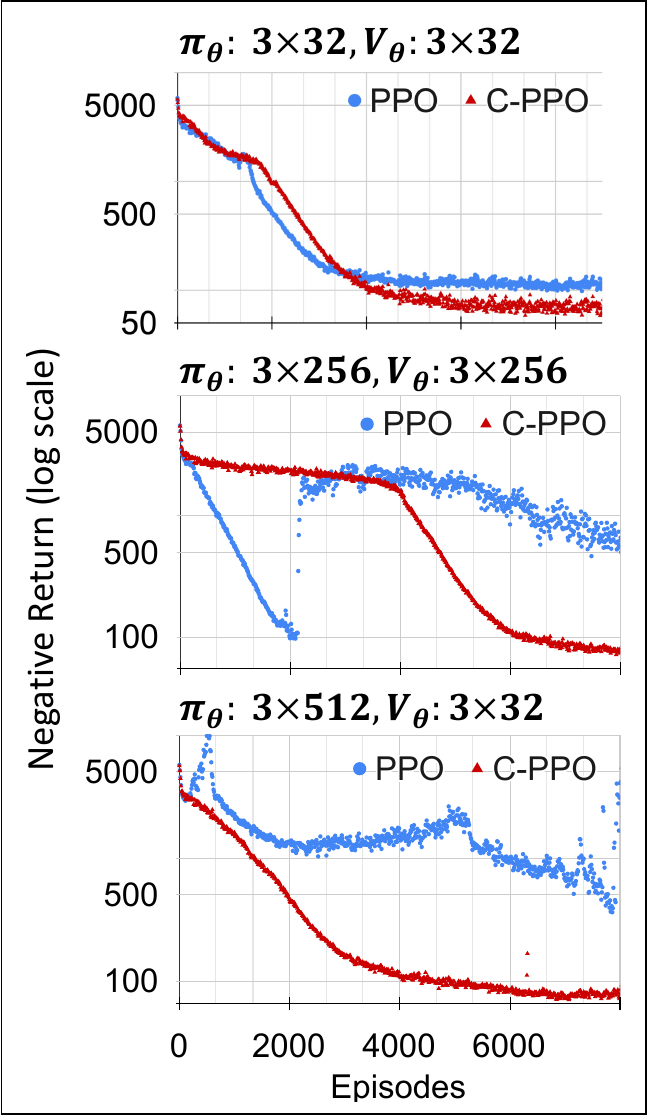}}\label{fig:overfitting}}
    \caption{Stability constraints (PPO with the soft terminal conditions vs. C(onstrained)-PPO): (a) (b) Histogram of absolute errors (average of the last 5 steps) in the generalization test. Policies at Episode 5000 are tested for $\bm\mu^{test}= \bm\mu^{n}+\Delta\bm \mu$ where $\Delta \bm\mu$ is sampled from the uniform distributions within $\pm 50\%$ $\tau_{i}^{n}$, $\pm 50\%$ $K_{j}^n$, and $\pm 100\%$ $K_{sur}^{n}$ ($K_{sur}^{test}>1$). The nominal parameters $\bm\mu^n$ are used to derive the stability constraints in C-PPO. PPO failed to maintain the control accuracy for the new parameter distribution, while C-PPO succeeded. (c) Large control networks ($3\times 256$ and $3\times 512$) can be used for learning, provided stability constraints on those networks, the constraints that can prevent overfitting to unstable solutions leading to the collapse of learning. }\label{fig:stability_ onstraints}
    \end{figure*}

We test if adding constraints enable learning with large networks, which suffers from overfitting. We gradually doubled the network sizes from $3\times 32$ until $3\times 512$ for both the control networks and value networks. The baseline is with $3\times 32$ for both (the top figure in Fig.~\ref{fig:overfitting}). 

PPO collapses with the network size at $3\times 256$ (the middle figure in Fig.~\ref{fig:overfitting}), with the initial convergence up to -100 and suddenly collapsed at Episode 2000. 

C-PPO shows unstable learning curves with $3\times 512$ for both value and control networks. We then decreased the size of value networks from $3\times 512$ to $3\times 32$ (control networks still $3\times 512$). C-PPO succeeded in learning the task again, while PPO still failed (the bottom figure in Fig.~\ref{fig:overfitting}). 

This suggests overfitting to the small fraction of unstable control corrupts the learning.

\section{Concluding Remarks}
We provide a new perspective about why RL struggles with generalization and robustness: given some dynamic parameter distributions, local optimal policies can perform unstable control for some dynamic parameters while achieving high rewards over the parameter distribution, because the stability boundaries do not necessarily align with the rewards landscape. Overfitting to such instabilities can deteriorate generalization and robustness. 

Our analysis and examples are from the perspective of control theory. The results open many questions for future study. For example, how often does this issue happen in RL? Can we leverage the implicit stability boundaries to estimate the bounds of generalization? Can we mathematically characterize the unstable control as adversarial samples in the sense of classification? Can we establish relation to the adversarial learning that also regularizes the input gradients?

%
%
%
%
%
%


\section*{Acknowledgments} This study was supported in part under the RIE2020 Industry Alignment
  Fund – Industry Collaboration Projects (IAF-ICP) Funding Initiative,
  as well as cash and in-kind contribution from the industry partner,
  HP Inc., through the HP-NTU Digital Manufacturing Corporate Lab.
\balance

\bibliographystyle{IEEEtran}
\bibliography{refs}




\end{document}